\documentclass[10pt,twocolumn,letterpaper]{article}
\usepackage[pagenumbers]{cvpr}
\usepackage{graphicx,amsmath,amssymb,booktabs,multirow,array,tabularx,xcolor}
\usepackage{microtype}
\usepackage{enumitem}
\setlist{nosep,leftmargin=*}
\newcommand{\pih}{\pi_{0.5}}
\newcommand{\sg}{\operatorname{sg}}
\newcommand{\BCE}{\operatorname{BCE}}
\newcommand{\para}[1]{\smallskip\noindent\textbf{#1}\enspace}
\newcommand{\figasset}[2][\linewidth]{\includegraphics[width=#1]{figures/#2.png}}
\definecolor{cvprblue}{rgb}{.21,.49,.74}
\usepackage[breaklinks,colorlinks,citecolor=cvprblue,linkcolor=cvprblue,urlcolor=cvprblue]{hyperref}
\usepackage{balance}
\def\confName{CVPR}
\def\confYear{2027}
\title{Learning to Move from and for Manipulation Policies: Spatializing Latent Action Representations for Goal-Pose-Free Mobile Manipulation}
\author{
Siyuan Zheng$^{1}$\quad Zhong Cao$^{2,*}$\quad Henry X. Liu$^{2,3,*}$\\[3pt]
{\small $^{1}$Department of Electrical and Computer Engineering, University of Michigan}\\
{\small $^{2}$Department of Civil and Environmental Engineering, University of Michigan}\\
{\small $^{3}$University of Michigan Transportation Research Institute}\\[3pt]
{\small $^{*}$Corresponding authors.}\\[4pt]
{\small Project website: \url{https://zsiyuan-eng.github.io/M4M/}}
}
\hypersetup{pdftitle={Learning to Move from and for Manipulation Policies: Spatializing Latent Action Representations for Goal-Pose-Free Mobile Manipulation},pdfauthor={Siyuan Zheng, Zhong Cao, Henry X. Liu}}
\makeatletter
\apptocmd{\@maketitle}{%
  \noindent\begin{minipage}{\textwidth}
    \centering
    \figasset[\textwidth]{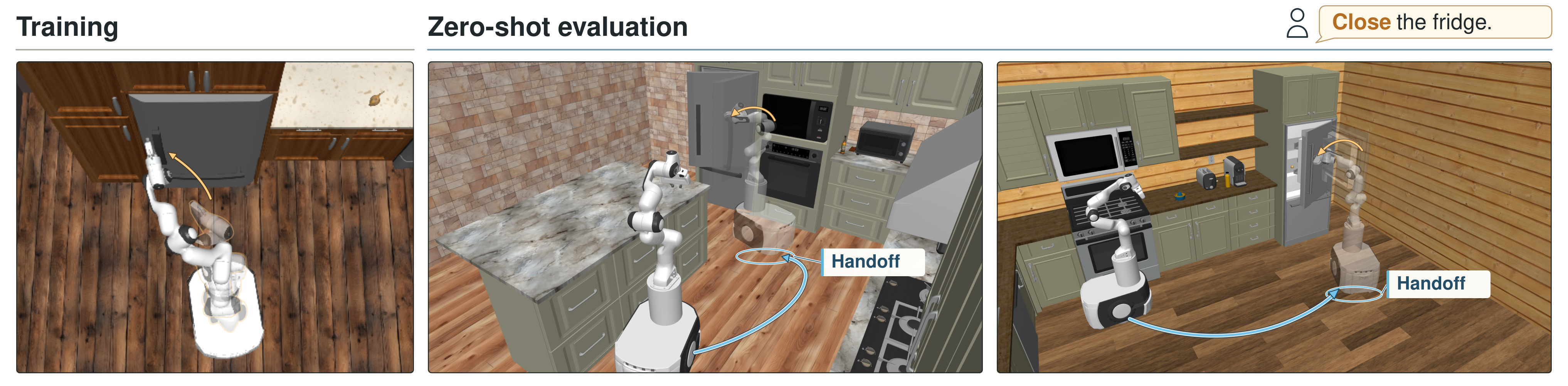}
    \captionsetup{hypcap=false}\captionof{figure}{\textbf{From training to zero-shot manipulation.}
    Left: three arm poses from a refrigerator-closing rollout in the training kitchen.
    Right: given the same instruction in two held-out kitchens, the robot repositions before handing off to the unchanged manipulation policy.
    Translucent robots show captured end poses; arrows indicate motion direction.}
    \label{fig:zero_shot}
  \end{minipage}\par\medskip
}{}{\PackageError{M4M}{Could not attach first-page teaser}{}}
\makeatother
\begin{document}
\maketitle
\raggedbottom
\begin{abstract}
The success of manipulation policies is closely tied to the spatial relationship between the robot and the target object, as well as the task being performed, making the ability to adjust the robot’s pose critical for successful execution. When the robot starts away from the task, such adjustments require coordinating navigation with local manipulation. Jointly training these capabilities incurs substantial computational and data costs, while navigating toward a predefined goal pose limits the flexibility needed to adapt to task-specific manipulation requirements. To address these challenges, we propose Learning to Move from and for Manipulation Policies (M4M), a framework that learns manipulation-aware mobility directly from pretrained manipulation policies. By leveraging the hidden representations of the policy’s action-generating network as dense supervision, together with a limited number of manipulation execution outcomes, we learn a spatially conditioned embedding that captures task-dependent manipulation success across candidate robot poses. From the robot’s current stereo observations, this representation estimates manipulation success at unvisited poses without rolling out the manipulation policy at each location. It is further integrated into navigation policy training, while a learned handoff policy decides when to transition from navigation to manipulation, enabling the robot to move and manipulate in a unified, task-aware manner. Across eight articulated-object tasks in unseen RoboCasa kitchens, M4M achieves 47.0\% mean task success, compared with 26.5\% for the strongest baseline and 19.2\% for direct manipulation execution, and refrigerator-closing experiments show consistent improvements across four frozen manipulation policies.
\end{abstract}
\begin{figure*}[t]
\centering
\figasset[\textwidth]{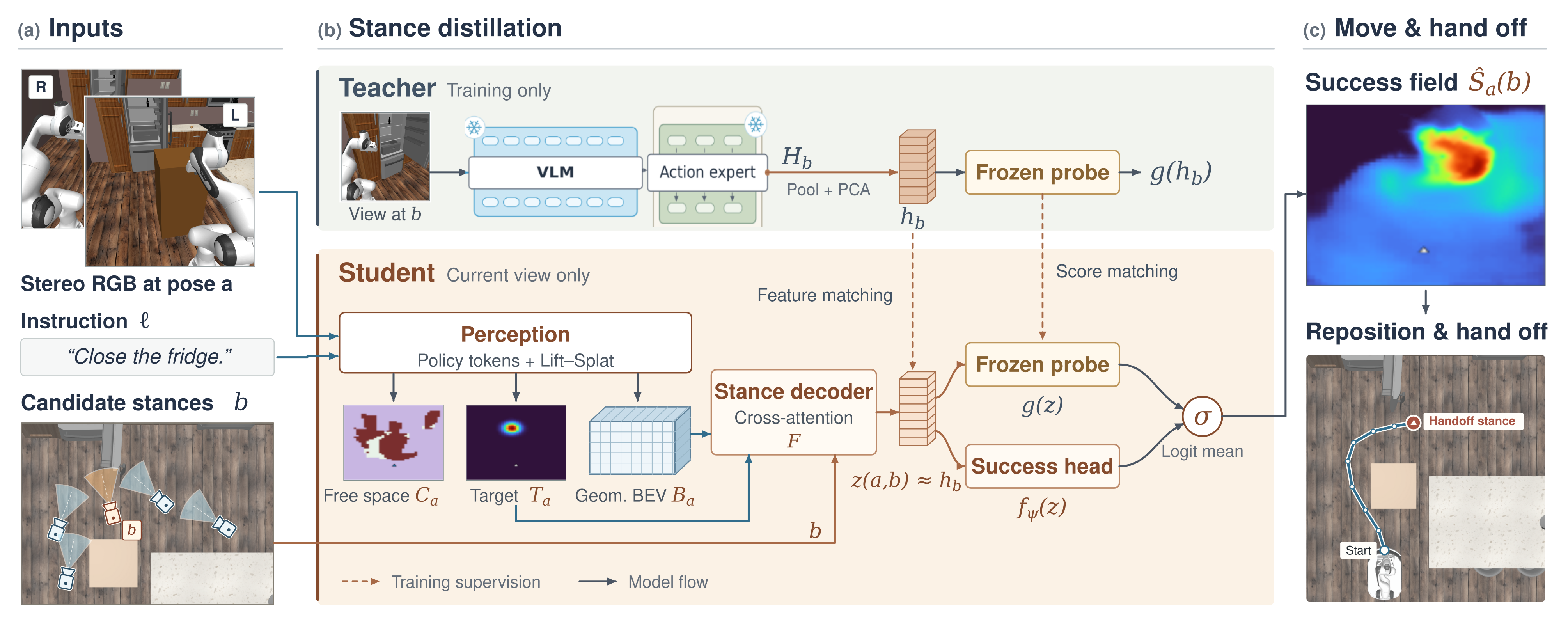}
\caption{\textbf{Success-guided approach and manipulation.} At training time, the frozen policy observes candidate stance $b$, supplying compressed action-expert features $h_b$ and the probe score $g(h_b)$. The student predicts $z(a,b)$ from stereo and language at the current pose $a$, conditioned on the candidate. Feature matching supervises $z$ and score matching supervises $g(z)$. The frozen probe and supervised SuccessHead form the success field $\hat S_a(b)$. A compact controller reads the success, target, and traversability fields to choose collision-checked moves and compare stopping with continuing before handoff. The success-field inset illustrates the predicted field used for movement and handoff.}
\label{fig:overview}
\end{figure*}

\section{Introduction}
A robot that can close a refrigerator from a favorable pose may still fail when it starts farther away. It must approach the appliance, avoid obstacles, and establish a distance and viewpoint from which its manipulation policy can complete the interaction. The handle may otherwise appear in an unfamiliar part of the image, or the base may leave insufficient space for the door's motion. Extending learned manipulation skills to mobile robots therefore requires coordinating the approach with the conditions under which those skills succeed~\citep{mobipi,n2m}.

\begin{figure*}[t]
\centering
\figasset[\textwidth]{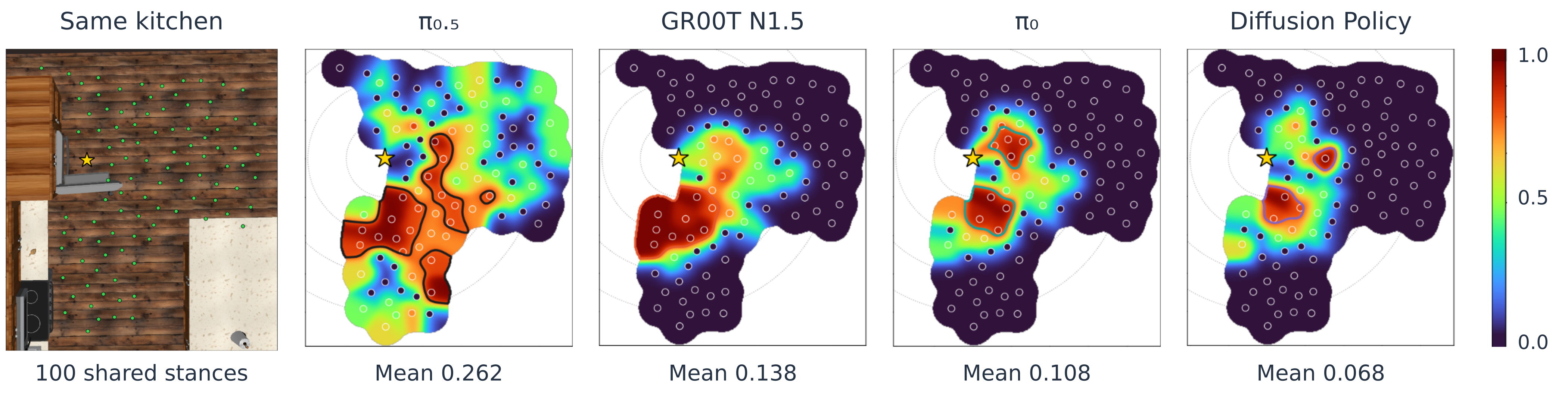}
\caption{\textbf{Manipulation stance preferences depend on the policy.} Four frozen policies---$\pih$~\citep{pi05}, GR00T N1.5~\citep{grootn15}, $\pi_0$~\citep{pi0}, and Diffusion Policy~\citep{diffusionpolicy}---are evaluated at the same 100 poses around a refrigerator. Circles mark sampled stances; the star marks the fixture. Color denotes empirical success from 0 to 1 in the shared coordinate frame. This diagnostic motivates policy-aware stance selection; its means are separate from the held-out multi-task benchmark.}
\label{fig:policy_stances}
\end{figure*}

Learning navigation and manipulation together from demonstrations requires data covering both behaviors and their coordination~\citep{mobipi}. Reusing an existing manipulation policy reduces what must be relearned, but leaves a central question: how should the robot move, and when should it hand off, before it has visited the candidate stances? Geometric reachability and target proximity do not fully capture a learned policy's preferences~\citep{n2m}. Figure~\ref{fig:policy_stances} illustrates this dependence: four manipulation policies evaluated at the same refrigerator poses exhibit different success landscapes.

We introduce \emph{M4M}, which learns both decisions from policy-derived success supervision. The policy's latent action representations at candidate stances are spatialized into a success field that a student predicts from the current view, so the robot can score stances before reaching them. A compact controller reads this field, together with the target and traversability maps, and decides at every step whether to keep moving or hand off, without committing to a target base pose. Goal-pose-free refers to this execution interface: candidate stances are scored, but no fixed navigation endpoint is selected.

Executing the manipulation policy at every training candidate is expensive. Instead, candidate-view action representations supply dense supervision, linked to success by limited execution outcomes. Cross-view distillation transfers this information to the current observation; privileged value iteration supervises movement and handoff. Deployment requires neither candidate-view rendering nor action-expert queries at candidate stances. Figure~\ref{fig:zero_shot} shows local approach and handoff in unseen kitchens, evaluated at 0.5--2.5\,m starts.

Our contributions are: (i) spatializing a frozen policy's latent action representations into a success field over unvisited stances; (ii) distilling policy-derived movement and handoff values into a controller that uses this field without explicit goal-pose selection; and (iii) evaluation on eight tasks in unseen kitchens, four-policy refrigerator experiments, ablations, representation comparisons, and a shared-budget study.

\section{Related Work}
\para{Language-guided object navigation.}
Open-vocabulary navigation grounds language in observations and maps to reach a named object or spatial goal. VLMaps fuses visual--language features into a spatial map~\citep{vlmaps}, ConceptGraphs builds open-vocabulary 3D scene graphs~\citep{conceptgraphs}, and CoW and VLFM combine target localization with exploration~\citep{cow,vlfm}. LM-Nav composes pretrained language, vision, and navigation models for instruction following~\citep{lmnav}. These approaches support finding and reaching targets; our success field addresses the subsequent choice of a stance suited to a particular manipulation policy. HomeRobot and OK-Robot integrate navigation with grasping and placement~\citep{homerobot,okrobot}. We focus on predicting how completion by an unchanged policy varies across candidate stances, including positions at similar target distances (Fig.~\ref{fig:policy_stances}).

\para{Mobile manipulation and stance selection.}
Inverse reachability maps select base poses from arm reachability and manipulability~\citep{reachabilityinversion}. N$^2$M$^2$ learns base motions that maintain manipulation feasibility~\citep{n2m2}, while ReLMoGen learns base and arm subgoals executed by a motion generator~\citep{relmogen}. MoManipVLA optimizes mobility and manipulation around VLA-predicted waypoints~\citep{momanipvla}, and MoTo selects docking points using VLM-generated interaction keypoints and feasibility objectives~\citep{moto}. Where2Act, VoxPoser, and SayCan score object interactions, spatial values, and skill feasibility~\citep{where2act,voxposer,saycan}; our field evaluates a fixed policy's success at unvisited base stances. Closest to our setting, Mobi-$\pi$ compares rendered candidate views with initial images from the policy's training demonstrations~\citep{mobipi}, whereas N2M learns pose preferences from rollouts~\citep{n2m}. We obtain dense supervision from the frozen policy's internal action states and predict candidate-specific success from the current observation, without rendering or visiting each candidate at deployment.

\para{Foundation policies.}
RT-1, RT-2, PaLM-E, OpenVLA, and the $\pi$ family connect visual and language knowledge with robot control and embodied reasoning~\citep{rt1,rt2,palme,openvla,pi0,pi05}; Octo learns a generalist manipulation policy from diverse robot trajectories~\citep{octo}. SpatialVLA, Transporter Networks, CLIPort, PerAct, and Act3D incorporate spatial structure into manipulation~\citep{spatialvla,transporter,cliport,peract,act3d}. These systems address the interaction itself; we retain the manipulation policy and choose the stance from which it operates.

\para{Which features predict success.}
RRL, R3M, and VC-1 show that pretrained visual representations support control and embodied tasks~\citep{rrl,r3m,vc1}. DINOv2 learns general-purpose visual features~\citep{dinov2}, while CLIP and SigLIP align images with language~\citep{clip,siglip}. RL Token builds a compact interface to VLA features for action refinement with online reinforcement learning~\citep{rltoken}. Motivated by these representation-learning approaches, we probe generic image features and successive stages of a frozen policy. The action-expert state ranks stances best across layouts (Sec.~\ref{sec:policy_feature_stages}).

\para{Cross-view distillation.}
Knowledge distillation and privileged-information learning transfer richer training-time information to simpler deployment inputs~\citep{distillation,privileged}; FitNets and Attention Transfer supervise intermediate features or attention patterns~\citep{fitnets,attentiontransfer}. Visual-policy distillation has also transferred knowledge across camera configurations~\citep{multicamkd}. Our asymmetry is spatial: the teacher observes candidate stance $b$ directly, whereas the student observes pose $a$ and receives $b$ as a query. Lift--Splat and Simple-BEV provide geometry for relating the views~\citep{lss,simplebev}. Feature matching and success-score preservation transfer policy-specific information to candidate stances that need not be observed online.
\section{Method}
\label{sec:method}
\begin{figure*}[t]
\centering
\figasset[\textwidth]{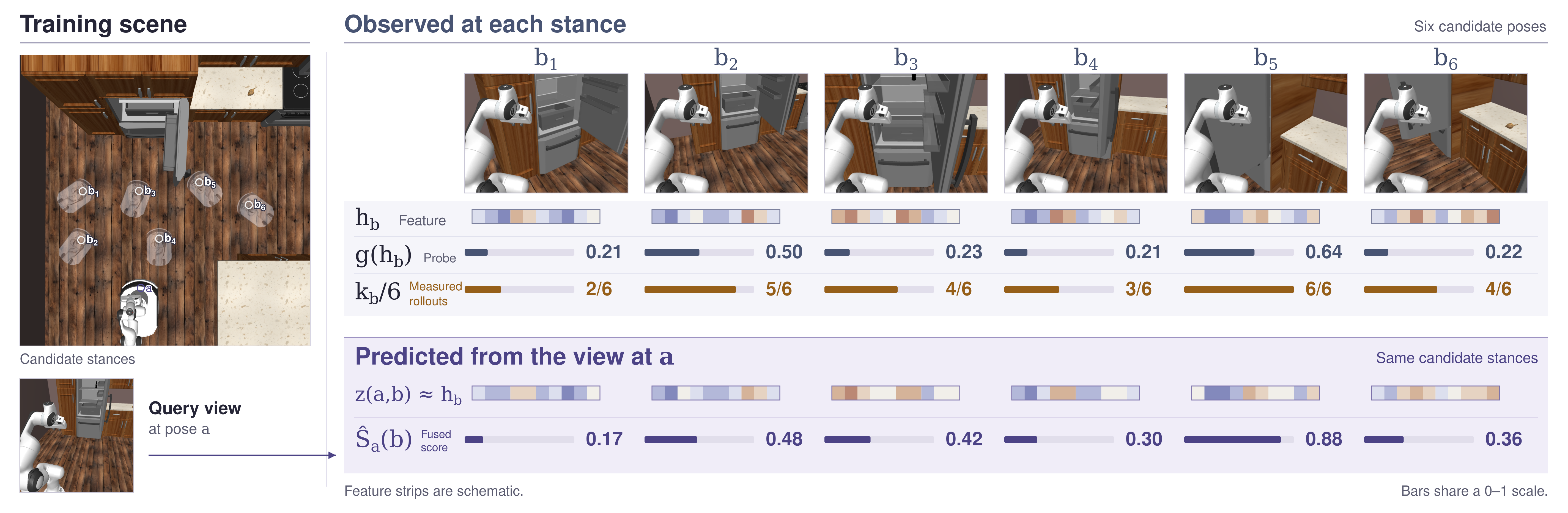}
\caption{\textbf{One current view predicts success across candidate stances.} Six candidate views in the training scene provide teacher features $h_b$, frozen-probe scores $g(h_b)$, and measured outcomes $k_b/6$. From the view at $a$, the student predicts $z(a,b)$ and the fused score $\hat S_a(b)$ for each of the same stances. The feature strips are schematic; the counts and scores are from the example.}
\label{fig:distillation}
\end{figure*}

\subsection{Problem and training resources}
At current base pose $a$, the robot receives calibrated stereo RGB $o_a=(I_a^L,I_a^R)$ and task instruction $\ell$. A candidate stance $b=(x,y,\theta)$ lies in the current base coordinate frame. For a fixed manipulation policy $\pi$, we seek a reachable stance from which executing $\pi$ completes the task. We train a predictor $\hat S_a(b)$ of this success and a lightweight controller that uses the predicted field to approach a suitable stance. The policy's weights remain unchanged throughout. Figure~\ref{fig:overview} traces the training and deployment paths.

We distinguish three training resources. Simulator geometry supplies target, depth, and traversability labels. Candidate observations expose policy representations without executing the proposed actions. A budgeted subset of candidates additionally has manipulation outcomes $(k_b,n_b)$: $k_b$ successful completions among $n_b$ executions. These outcome records fit the probe and supervise the student. Navigation episodes train the movement and handoff controller. In the budget comparison, these episodes are counted together with stance-label executions in the total rollout budget; dense feature extraction requires no manipulation execution.

\subsection{Geometry and semantic grounding}
An ImageNet-initialized ResNet-18 encodes the stereo images, predicts image-depth distributions, and lifts features into BEV using Lift--Splat~\citep{lss}. Its geometric feature map $B_a$ represents local spatial structure. This backbone is trained with simulator geometric supervision and fixed during stance distillation; it does not consume manipulation-success labels.

A separate semantic path reuses the frozen policy's visual tokens, before the language-model transformer. Following geometric unprojection~\citep{simplebev}, BEV columns at sampled heights are projected into both cameras, and their corresponding visual features are sampled and fused into $f_a(x)$. The policy's language model encodes the instruction as $e_\ell$. Learned projections match the spatial content to this task query:
\begin{equation}
 T_a(x)=\sigma\!\left(\langle W_vf_a(x),W_\ell e_\ell\rangle/\tau+b_T\right).
 \label{eq:target}
\end{equation}
The target head is supervised by the task object's position when visible and by an all-zero map otherwise. A separate traversability head supplies the semantic map $C_a$, including blocked-cell probabilities; depth-derived visibility marks unobserved cells as unknown. The controller uses these probabilities and the inverse unknown mask as separate inputs.

The two paths serve different roles. $B_a$ and $T_a$ condition stance decoder $F$, while $C_a$ informs navigation and swept-path collision checking. Language conditions $T_a$ through matching and $B_a$ through FiLM; the pose decoder receives these conditioned maps rather than raw text. The final manipulation policy also receives the instruction. Perception modules use geometry labels and are frozen before the stance student is fitted.

\subsection{Candidate-view teacher and success probe}
Visual tokens encode the observation, and the language-model prefix fuses it with the instruction. We take the teacher from the next stage, where the action expert computes motion conditioned on that context. Section~\ref{sec:policy_feature_stages} compares these stages with the decoded actions.

For a training anchor $a$, we observe the scene at candidate stance $b$ and run the unchanged manipulation policy on its standard observation. We extract the final action-expert hidden states $H_b$ before action projection. Pooling, training-set standardization, and principal component analysis produce a compact target with $d=128$
\begin{equation}
 h_b=P_H\operatorname{Std}_H(\operatorname{Pool}(H_b))\in\mathbb R^d.
 \label{eq:teacher}
\end{equation}
Preprocessing is fitted only on training candidates. The teacher forward pass computes an action sequence but does not execute it. It therefore supplies targets at candidates lacking outcome labels, without additional manipulation rollouts.

A logistic probe associates these states with observed task completion:
\begin{equation}
 g(h)=\sigma(w^\top h+c),\qquad \ell_g(h)=w^\top h+c.
\end{equation}
Let the binomial negative log likelihood, up to a parameter-independent constant, be
\begin{equation}
 \mathcal B(k,n;p)=-k\log p-(n-k)\log(1-p).
 \label{eq:binomial}
\end{equation}
We first fit $w,c$ using $\mathcal B(k_b,n_b;g(h_b))$ on the budgeted labeled candidates $\mathcal L$. We then freeze the probe. It scores every feature-paired training candidate in $\mathcal U$, supplying dense soft targets $g(h_b)$.

\subsection{Current-view stance distillation}
The student predicts a candidate representation from the current view,
\begin{equation}
 z_\theta(a,b)=F_\theta(B_a,T_a,b)\in\mathbb R^d.
 \label{eq:student}
\end{equation}
Figure~\ref{fig:distillation} shows this prediction from a single current observation for six candidate stances. A query embeds position, target-facing heading, and locally sampled BEV features. Cross-attention reads pooled geometric tokens and a target-weighted token. This spatial grounding lets a distant observation predict a feature associated with a different candidate view. Candidates have no self-attention between them, so a query's meaning does not depend on the other candidates in the batch.

Feature matching constrains the predicted representation, while score matching preserves the teacher's success-relevant direction. With stop-gradient $\sg$ and binary cross entropy $\BCE$, the dense losses are
\begin{align}
 \mathcal L_H&=\mathbb E_{(a,b)\in\mathcal U}\frac1d\|z_\theta(a,b)-\sg(h_b)\|_2^2,\\
 \mathcal L_g&=\mathbb E_{(a,b)\in\mathcal U}\BCE\big(g(z_\theta(a,b)),\sg(g(h_b))\big).
 \label{eq:dense}
\end{align}
Because the probe is linear before its sigmoid, score preservation emphasizes the projection $w^\top z$, rather than requiring every teacher coordinate to be equally useful for success.

SuccessHead is an independent $128\to64\to1$ MLP trained from scratch. Its logit $f_\psi(z)$ is supervised only by the observed rollout counts on $\mathcal L$, with $p_\psi(z)=\sigma(f_\psi(z))$. The complete student objective is
\begin{equation}
 \mathcal L=\mathcal L_H+\mathcal L_g+
 \mathbb E_{(a,b)\in\mathcal L}\mathcal B(k_b,n_b;p_\psi(z_\theta(a,b))).
 \label{eq:total}
\end{equation}
The reported model uses unit weights for all three terms. Only $F_\theta$ and SuccessHead are updated; the policy, geometric modules, preprocessing, and probe remain fixed. A frozen probe fixes parameters but still differentiates with respect to its input, so $\mathcal L_g$ updates the student.

The frozen probe is fitted on teacher features; dense score matching trains the student to preserve the probe's success-relevant projection. SuccessHead instead learns from observed outcomes on the student's own predicted features, so it can adapt to systematic feature errors but has fewer supervision targets. At deployment, averaging the logits combines these complementary signals:

\begin{equation}
 \hat S_a(b)=\sigma\!\left(\tfrac12[\ell_g(z_\theta(a,b))+f_\psi(z_\theta(a,b))]\right).
 \label{eq:fusion}
\end{equation}
Both readouts operate on predicted $z$, so candidate evaluation does not query the action expert.

\subsection{Learned repositioning and value-based handoff}
A CNN takes predicted fields $S_t,T_t,C_t\odot O_t,O_t$, normalized base-frame coordinates and elapsed time as input $x_t$, predicting movement directions and stopping/continuation values.

During training, simulator traversability $C^*$ and target positions support a dense field $S^*(b)=g(h_b)$: target-facing stances on a 0.1\,m grid supply rendered frozen-policy features, without manipulation execution. Scores use the outcome-fitted probe. Convolution $\widetilde S=S^*\circledast K$ with a normalized empirical arrival-error kernel (approximately 0.1\,m) favors robust success regions.

Value iteration considers stopping and 16 translation directions feasible under $C^*$. With step cost $c>0$,
\begin{equation}
 \begin{aligned}
 Q^*(q,a)&=V^*(q+\Delta_a)-c,\\
 V^*(q)&=\max\{\widetilde S(q),U^*(q)\},
 \end{aligned}
 \label{eq:controller_teacher}
\end{equation}
Here $U^*(q)=\max_{a\in\mathcal A(q)}Q^*(q,a)$, with feasible moves $\mathcal A(q)$ and displacements $\Delta_a$. This maximizes success score minus travel cost in the grid model. The direction head fits $\operatorname{softmax}_{a\in\mathcal A(q)}(Q^*(q,a)/\eta)$ with temperature $\eta$; the value heads fit $\widetilde S(q)$ and $U^*(q)$, respectively. Student rollouts supply inputs; true poses index these targets without manipulation execution. Later-round outcomes recalibrate stopping values at actual handoffs.

Deployment uses predicted fields and the distilled stopping rule
\begin{equation}
 V_{\rm stop}(x_t)\geq V_{\rm cont}(x_t).
 \label{eq:gate}
\end{equation}
Otherwise, collision checking selects a feasible 0.15\,m move by direction rank; the base faces the target and re-observes. Manipulation and success-predictor weights remain frozen. Details are in the supplement.
\section{Experiments}
\label{sec:experiments}
We ask whether policy-feature distillation improves final manipulation completion, which stage of the frozen policy yields the most transferable stance ranking, and how the improvement changes with the total rollout budget.

\begin{figure*}[t]
\centering
\figasset[\textwidth]{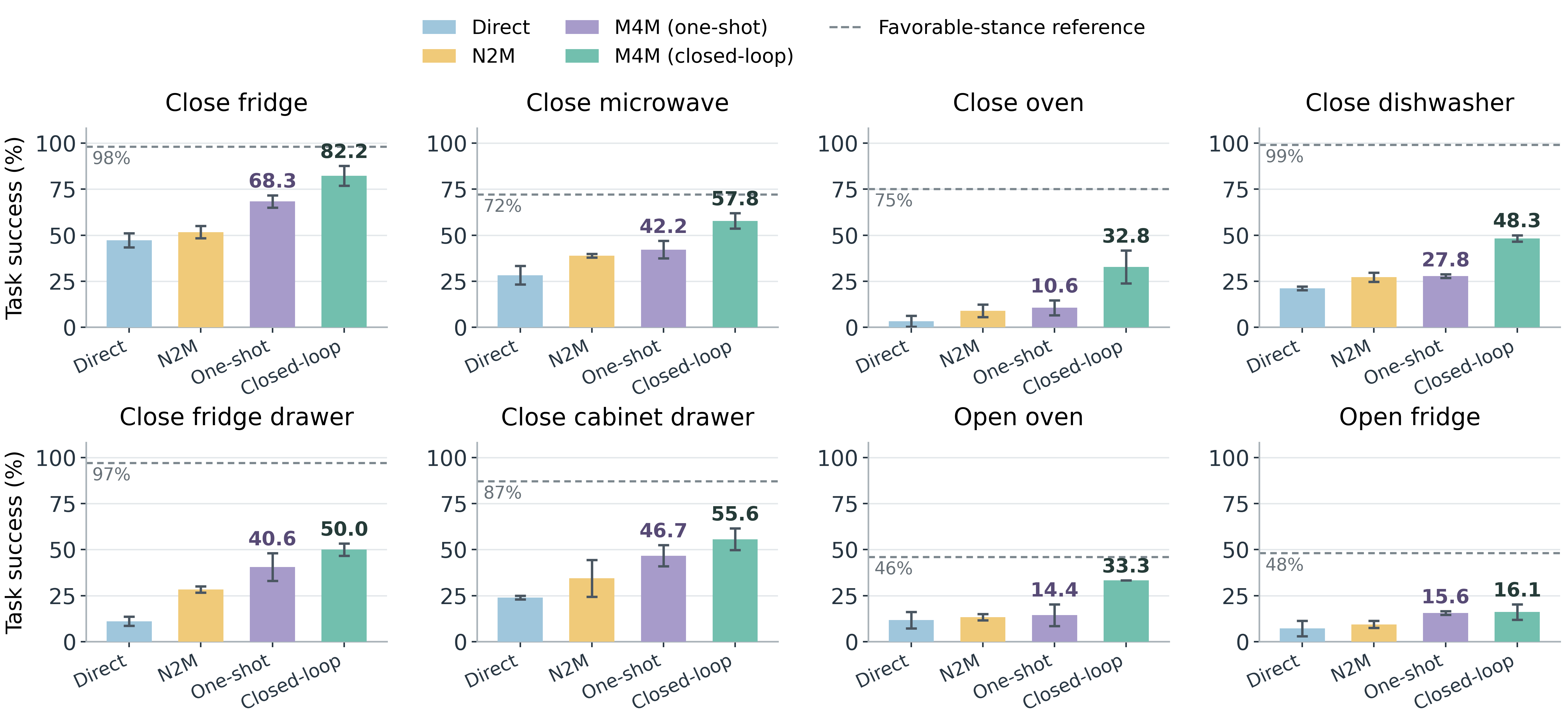}
\caption{\textbf{Eight-task completion on unseen kitchens.} Each task uses two held-out kitchens and 180 episodes per method; error bars show SD. N2M and M4M (one-shot) predict once from the initial view, followed by placement and manipulation. M4M (closed-loop) moves and re-observes before handoff. Dashed lines mark separately measured favorable-stance references.}
\label{fig:multitask}
\end{figure*}

\subsection{Evaluation protocol}
\para{Tasks and held-out kitchens.}
We evaluate eight articulated-object tasks in RoboCasa~\citep{robocasa}: closing a refrigerator, microwave, oven, dishwasher, refrigerator drawer, and cabinet drawer, and opening an oven and a refrigerator. Every task uses two held-out kitchen layouts and starting distances from 0.5 to 2.5\,m. Evaluation kitchens are absent from training. The eight-task benchmark trains a separate predictor for each task and uses the same frozen $\pih$ manipulation checkpoint throughout. The main setting uses a total budget of 1,000 rollouts: 800 for stance-label supervision and 200 refrigerator-closing rollouts for the navigation and handoff controller, which transfers unchanged to the other seven tasks. This transfer concerns the controller; each task retains its own distilled success field. Teacher feature extraction and geometry supervision require no additional manipulation outcomes.

Each task contributes 180 evaluation episodes per method across its two kitchens. Independent experiments vary environment initialization across layouts and starting distances; reported uncertainty is the standard deviation across these experiments. Success means that the frozen policy completes the requested task after handoff; navigation arrival alone is not counted as completion.

\para{Compared systems.}
\emph{Direct $\pih$} executes from the starting pose without a stance-selection module. \emph{N2M+$\pih$} follows the rollout-labeling protocol~\citep{n2m}: stances are randomly sampled near the target, with one frozen-policy execution per sampled stance. At test time it predicts from the initial observation, followed by placement and execution. \emph{M4M} repositions the base in simulation using repeated observations and uses the learned movement policy and value comparison in Sec.~\ref{sec:method}. M4M (one-shot) uses the same initial-view prediction, placement, and execution protocol as N2M across all eight tasks. Total budgets include stance-label collection and controller training. We do not directly evaluate Mobi-$\pi$: its original scoring requires policy-training demonstration images and test-scene reconstruction~\citep{mobipi}, neither of which is provided in our evaluation setting.

\subsection{Eight-task manipulation success}
Figure~\ref{fig:multitask} shows higher mean completion for our method on all eight tasks. Equal-task macro success is \textbf{47.0\%}, compared with 26.5\% for N2M and 19.2\% for direct execution. Relative to N2M, the absolute gains range from 6.7 to 30.5 percentage points.

\begin{table*}[t]
\centering\small
\caption{\textbf{Repositioning analysis on two unseen fridge kitchens.} (a) Stance-scoring and navigation ablations: three groups of 60 episodes, 90 per layout, at 0.5--2.5\,m. (b) Box obstacle: three groups of 24 episodes, 36 per layout, at 2.0/2.5\,m with obstacle geometry available only during training. Values are success rates (\%, mean $\pm$ SD across groups).}
\label{tab:ablation}\label{tab:obstacle}
\begin{minipage}[t]{.53\textwidth}
\centering\textbf{(a) Refrigerator ablations}\par\smallskip
\begin{tabular*}{\linewidth}{@{\extracolsep{\fill}}lr@{}}\toprule
Condition & Success (\%)\\\midrule
Empirical standoff (1.3\,m) & $51.7\pm6.0$\\
Distilled probe only & $69.4\pm5.1$\\
Outcome-only head & $75.6\pm7.9$\\
Dijkstra & $72.8\pm6.0$\\
\textbf{Full model} & $\mathbf{82.2\pm5.4}$\\\bottomrule
\end{tabular*}
\end{minipage}\hfill
\begin{minipage}[t]{.43\textwidth}
\centering\textbf{(b) Box obstacle}\par\smallskip
\begin{tabular*}{\linewidth}{@{\extracolsep{\fill}}lr@{}}\toprule
Method & Success (\%)\\\midrule
Direct $\pih$ & $27.8\pm2.4$\\
N2M & $29.2\pm0.0$\\
\textbf{M4M} & $\mathbf{47.2\pm2.4}$\\\bottomrule
\end{tabular*}
\end{minipage}
\end{table*}

\begin{table*}[!t]
\centering\footnotesize
\setlength{\tabcolsep}{5pt}
\caption{\textbf{Cross-layout selection regret} (lower is better). Probes train on one kitchen and evaluate on three unseen layouts. Values are mean $\pm$ SD over five label-subset repeats; bold marks the lowest mean. Supplementary Table S6 also reports within-kitchen results.}
\label{tab:features_full}
\begin{tabular*}{\textwidth}{@{\extracolsep{\fill}}lccccc@{}}\toprule
Labels & ResNet-18 & SigLIP & Prefix & $H$ & Action\\\midrule
5\% & $0.274\pm 0.076$ & $0.240\pm 0.091$ & $0.211\pm 0.045$ & $\mathbf{0.174}\pm 0.015$ & $0.275\pm 0.051$\\
20\% & $0.293\pm 0.034$ & $0.248\pm 0.032$ & $0.177\pm 0.014$ & $\mathbf{0.148}\pm 0.018$ & $0.258\pm 0.045$\\
50\% & $0.262\pm 0.007$ & $0.228\pm 0.023$ & $0.176\pm 0.019$ & $\mathbf{0.151}\pm 0.015$ & $0.216\pm 0.009$\\
100\% & $0.264\pm 0.003$ & $0.210\pm 0.005$ & $0.176\pm 0.011$ & $\mathbf{0.158}\pm 0.003$ & $0.214\pm 0.010$\\
\bottomrule
\end{tabular*}
\end{table*}

\para{Task dependence and favorable initialization.}
Task dependence remains substantial. Refrigerator-drawer completion rises from 11.1\% for direct execution to 50.0\%, versus 28.3\% for N2M. Opening remains difficult: our success is 33.3\% for the oven and 16.1\% for the fridge, versus N2M's 13.3\% and 9.4\%. The controller is transferred from refrigerator closing in all three cases. Target distance alone does not specify a useful stance.

Dashed references in Fig.~\ref{fig:multitask} report favorable-stance completion without navigation; lower rates on the two opening tasks help contextualize their end-to-end results.

\para{Matched one-shot stance selection.}
With navigation removed, our one-shot predictor reaches 33.3\% macro success versus N2M's 26.5\%, improving the mean on all eight tasks. Refrigerator closing reaches $68.3\pm3.3$\% versus $51.7\pm3.3$\%. Closed-loop execution reaches 47.0\% macro success, a further 13.7-point gain over our one-shot setting. Rollout videos are provided in the supplementary material.

\subsection{Which parts improve repositioning?}

Table~\ref{tab:ablation}(a) compares five conditions on the same fridge starts. Empirical standoff stops approximately 1.3\,m from the refrigerator center, a distance chosen from the favorable operating region in the pose-success diagnostic (Fig.~\ref{fig:policy_stances}). Evaluated in the same two held-out kitchens, it trails the full model by 30.5 points: an empirically favorable distance alone does not determine a policy-compatible stance. Distilled probe only deploys $g(z)$ as its sole score. Outcome-only head retains the decoder and SuccessHead, trains only with the outcome NLL (without $\mathcal L_H$ or $\mathcal L_g$), and deploys $\sigma(f_\psi(z))$. Full model uses all three losses and the logit fusion in Eq.~\eqref{eq:fusion}. Full model exceeds Outcome-only head by 6.6 percentage points. These end-to-end comparisons vary training and readout jointly; Sec.~\ref{sec:loss_ablation} isolates the losses offline.

With shared fields, collision checks, and handoff gate, Dijkstra movement~\citep{dijkstra1959} reaches $72.8\pm6.0$\%, versus $82.2\pm5.4$\% for the learned controller.

\para{Obstacle-aware execution.}
For Table~\ref{tab:obstacle}(b), we place a box along the approach path and evaluate starts at 2.0 and 2.5\,m. Box geometry is available only during training; at test time, traversability is predicted from stereo observations without access to the box position or a prebuilt map. Over three experiments of 24 episodes, success is $47.2\pm2.4$\% for our system, $29.2\pm0.0$\% for N2M, and $27.8\pm2.4$\% for direct execution. The learned controller improves completion in the obstructed setting by using the success field together with semantic traversability and swept-path collision checks.

\subsection{Policy features as success supervision}
\label{sec:policy_feature_stages}
We compare five representations using matched labels, the same logistic-probe protocol, and PCA-64. ImageNet ResNet-18 supplies generic visual features. The other four representations come from the same forward pass of frozen $\pih$, in computation order: SigLIP visual tokens before language fusion, the fused image--instruction prefix before the action expert, the final action-expert hidden state $H$ before action projection, and the decoded 50-step action chunk. Selection regret is the empirical success gap between the best available stance and the stance chosen by the probe.

On unseen layouts, regret falls from SigLIP to Prefix to $H$ at every label budget (Table~\ref{tab:features_full}). With 20\% of the labels, the means are 0.248, 0.177, and 0.148, respectively, versus 0.293 for ResNet-18. The Prefix lowers mean regret by 0.071 relative to SigLIP; $H$ lowers it by a further 0.029 (16\% relative to Prefix). The $H$--Prefix gap remains 0.018 even at full labels. Its mean regret with 5\% of the labels (0.174) is lower than that of every other representation at 100\%, although the gap to Prefix (0.176) is small.

Decoding the motion reverses the trend: at 100\% of the labels, Action (0.214) is close to SigLIP (0.210) and well above $H$ (0.158). The same reversal holds at every budget: the probe on decoded actions ranks stances worse than the probe on $H$. Within the training kitchen, Action leads at 5\%, $H$ and Prefix are nearly tied at 20\%, and $H$ leads at 50\% and 100\% (Table S6); the consistent advantage appears when the layout changes.

\subsection{Loss ablation}
\label{sec:loss_ablation}
Table~\ref{tab:loss_ablation} evaluates the current-view student offline in unseen kitchens at a total budget of 1,000 rollouts over five seeds. This regret evaluates the current-view student and is not comparable to the candidate-view probes in Table~\ref{tab:features_full}. All settings use the same SuccessHead readout. Outcome supervision is retained; feature and score matching are varied independently. The combined losses give the lowest mean regret (0.312 versus 0.414) and Brier score (0.303 versus 0.379). Score matching alone improves mean regret more than feature matching alone, while feature matching gives the lower Brier score.

\begin{table}[t]
\centering\footnotesize
\setlength{\tabcolsep}{3pt}
\caption{\textbf{Offline loss ablation} (1,000-rollout budget, unseen kitchens; mean $\pm$ SD over five seeds). $H$ and $g$ denote feature and score matching; all rows include outcome supervision and use SuccessHead. Lower is better.}
\label{tab:loss_ablation}
\begin{tabular*}{\columnwidth}{@{\extracolsep{\fill}}lcccc@{}}\toprule
Setting & $H$ & $g$ & Regret $\downarrow$ & Brier $\downarrow$\\\midrule
A & -- & -- & $0.414\pm0.030$ & $0.379\pm0.019$\\
B & $\checkmark$ & -- & $0.412\pm0.038$ & $0.364\pm0.026$\\
C & -- & $\checkmark$ & $0.372\pm0.041$ & $0.373\pm0.025$\\
D & $\checkmark$ & $\checkmark$ & $\mathbf{0.312}\pm0.057$ & $\mathbf{0.303}\pm0.026$\\\bottomrule
\end{tabular*}
\end{table}

\subsection{Total rollout budget}
\begin{figure}[t]
\centering
\figasset{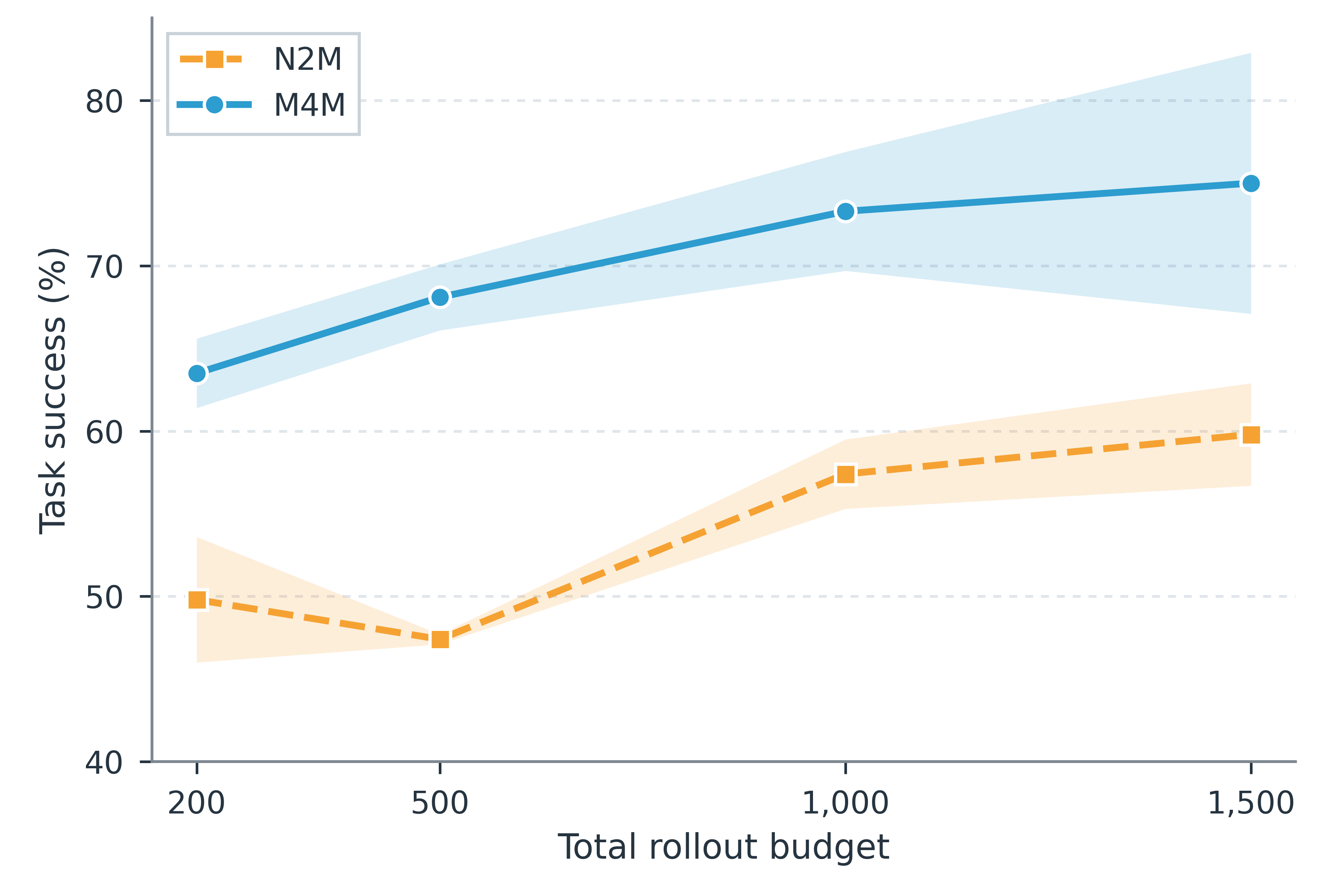}
\caption{\textbf{Matched total budgets of 200--1,500 rollouts.} Lines show mean completion; shading is one SD across three evaluation blocks. Budgets include both stance-label collection and controller training (100 controller rollouts at budget 200; 200 at each larger budget).}
\label{fig:budget}
\end{figure}

Figure~\ref{fig:budget} compares refrigerator closing at matched total rollout budgets, including controller training. Of 200 rollouts, 100 supervise the probe and stance predictor and 100 train the controller. At budgets of 500/1,000/1,500, 200 train the controller and 300/800/1,300 supply stance labels. The controller is fixed across these larger budgets. Each budget uses three sampled stance-label sets. The three training-subset draws in this study are independent of those used in the main benchmark. Evaluation uses three blocks on two held-out kitchens, with 60 episodes per trained model per block.

Across these total budgets, completion rises from $63.5\pm2.1$\% at a total budget of 200 rollouts to $68.1\pm2.0$\% at 500, $73.3\pm3.6$\% at 1,000, and $75.0\pm7.9$\% at 1,500. Gains over N2M range from 13.7 to 20.7 percentage points.

\subsection{Repositioning across manipulation policies}
\label{sec:across_policies}
We repeat the refrigerator-closing test on the same two held-out kitchens with $\pi_0$, GR00T N1.5, and Diffusion Policy in addition to $\pih$. Each row of Table~\ref{tab:across_policies} holds the frozen policy fixed. Across all four policies, the teacher reads final hidden states on action tokens before output projection (Appendix~A). Each manipulation policy has its own distilled predictor and separately trained navigation controller.

\begin{table}[!ht]
\centering\footnotesize
\setlength{\tabcolsep}{3.6pt}
\caption{\textbf{Refrigerator closing across frozen policies.} Success (\%, mean $\pm$ SD) uses the same held-out kitchens and 0.5--2.5\,m starts as the $\pih$ test. Direct executes from the initial pose; bold marks the best mean.}
\label{tab:across_policies}
\begin{tabular*}{\columnwidth}{@{\extracolsep{\fill}}lccc@{}}\toprule
Policy & M4M & N2M & Direct \\\midrule
$\pih$ & $\mathbf{82.2}\pm 5.4$ & $51.7\pm 3.3$ & $47.2\pm 3.8$ \\
$\pi_0$ & $\mathbf{55.0}\pm 6.0$ & $14.4\pm 1.9$ & $5.0\pm 1.7$ \\
GR00T N1.5 & $\mathbf{55.6}\pm 9.8$ & $16.1\pm 3.8$ & $4.4\pm 2.5$ \\
Diffusion Policy & $\mathbf{23.3}\pm 5.0$ & $12.2\pm 4.2$ & $3.3\pm 1.7$ \\\bottomrule
\end{tabular*}
\end{table}

Our method has the highest mean success for all four policies. Gains over N2M range from 11.1 percentage points for Diffusion Policy to 40.6 for $\pi_0$. Direct success is similarly low for GR00T N1.5 (4.4\%) and Diffusion Policy (3.3\%), yet their success after repositioning differs substantially (55.6\% versus 23.3\%). A policy's initial-pose performance therefore gives little indication of how much suitable repositioning can recover.

\section{Limitations}
Evaluation covers simulated kitchens and 0.5--2.5\,m starts; predictors are task-specific and cross-policy tests cover refrigerator closing. Training uses privileged geometry and rendered candidate views; deployment uses stereo observations and task instructions without test-scene reconstruction. Real-world teacher supervision would require candidate-view capture or scene reconstruction; robustness to rendering errors is untested. Probe, perception, and arrival-model errors may limit transfer. Physical robots, longer-range navigation, and moving obstacles remain untested.

\section{Conclusion}
M4M spatializes a frozen policy's action-expert representations into a success field over unvisited stances. Dense candidate-view supervision and limited execution outcomes train a current-view predictor; a controller uses the field with target and traversability maps to move and hand off without selecting a fixed goal pose.

Across eight tasks in unseen RoboCasa kitchens, M4M reaches 47.0\% mean success, versus 26.5\% for N2M and 19.2\% for direct execution. Its one-shot variant reaches 33.3\%, separating stance-prediction gains from closed-loop execution. Refrigerator-closing gains extend across four frozen policies, and the controller transfers unchanged to seven other tasks. Representation and loss ablations support policy-feature supervision.

Future work will test physical robots and improve uncertainty handling, calibration, and transfer. Code and videos are available on the project website: \url{https://zsiyuan-eng.github.io/M4M/}.
\label{main-body-end}

\section*{LLM Usage Statement}
We used Claude and Claude Code (Anthropic), and ChatGPT (OpenAI), for assistance with revising text, reviewing clarity and organization, suggesting figure layouts and LaTeX formatting, and debugging code. The authors designed the research, conducted the experiments, and made the conclusions. The authors are responsible for verifying all scientific claims, results, and code and for the final content of the submission.

\nobalance
{\small\bibliographystyle{ieeenat_fullname}\bibliography{references}}
\end{document}